\documentclass[preprint, letterpaper, 10pt, conference]{ieeeconf}

\IEEEoverridecommandlockouts

\usepackage{cite}
\usepackage{amsmath,amssymb,amsfonts}
\usepackage{algorithmic}
\usepackage{graphicx}
\usepackage{textcomp}
\usepackage{xcolor}
\def\BibTeX{{\rm B\kern-.05em{\sc i\kern-.025em b}\kern-.08em
    T\kern-.1667em\lower.7ex\hbox{E}\kern-.125emX}}

\usepackage{acro}
\usepackage{siunitx}
\usepackage{bm}
\usepackage{nicefrac}
\usepackage{algorithm}
\usepackage{pict2e}
\usepackage{pgfplots}
\pgfplotsset{compat=newest}

\acsetup{patch/maketitle=false}

\DeclareAcronym{ODE}{
    short = ODE,
    long = ordinary differential equation
}
\DeclareAcronym{LPV}{
    short = LPV,
    long =  linear parameter-varying
}

\DeclareAcronym{PI}{
    short = PI,
    long = proportional integral
}

\DeclareAcronym{FOPDT}{
    short = FOPDT,
    long = first order plus dead time
}

\DeclareAcronym{LMI}{
    short = LMI,
    long = linear matrix inequalities
}

\DeclareAcronym{LSTM}{
    short = LSTM,
    long = long short-term memory
}

\DeclareAcronym{RL}{
    short = RL,
    long = reinforcement learning
}
\DeclareAcronym{DRL}{
    short = DRL,
    long = deep reinforcement learning
}
\DeclareAcronym{MDP}{
    short = MDP,
    long = Markov decision process
}
\DeclareAcronym{SAC}{
    short = SAC,
    long = soft actor critic
}
\DeclareAcronym{TQC}{
    short = TQC,
    long = truncated quantile critic
}
\DeclareAcronym{DroQ}{
    short = DroQ,
    long = dropout Q-function
}

\DeclareAcronym{ReACT}{
    short = ReACT,
    long = regularized actor and critic TQC
}

\DeclareAcronym{BSO}{
    short = BSG-object,
    long = B-spline geometry object
}

\DeclareAcronym{BSG}{
    short = BSG,
    long = B-spline geometry,
    long-plural-form = B-spline geometries
}

\DeclareAcronym{NURBS}{
    short = NURBS,
    long = non-uniform rational B-splines,
    long-plural-form = non-uniform rational B-splines
}

\DeclareAcronym{CP}{
    short = CP,
    long = control point
}

\DeclareAcronym{EMA}{
    short = EMA,
    long = exponential moving average
}

\newcommand{\matlab}{\textsc{MATLAB}\textsuperscript{\textregistered} }
\newcommand{\simulink}{\textsc{Simulink}\textsuperscript{\textregistered} }

\begin{document}

\title{
\vspace{-3.5em}\footnotesize This work has been accepted at the IEEE for publication.\\
Copyright may be transferred without notice, after which this version will no longer be accessible.\\\vspace{5.25em}%
\LARGE \bf
ReACT: Reinforcement Learning for \\Controller Parametrization using B-Spline Geometries
}

\author{Thomas Rudolf\,$^{*1}$, Daniel Flögel\,$^{*1}$, Tobias Schürmann\,$^{1}$, Simon Süß\,$^{1}$, Stefan Schwab\,$^{1}$, and Sören Hohmann\,$^{2}$%
\thanks{$^*$\,These authors contributed equally: Thomas Rudolf, Daniel Flögel}%
\thanks{$^{1}$\,Thomas Rudolf, Daniel Flögel, Tobias Schürmann, Simon Süß, and Stefan Schwab are with the Embedded Systems and Sensors Engineering department of the FZI Research Center for Information Technology,
        76131 Karlsruhe, Germany
        \texttt{\{rudolf, floegel\}@fzi.de}}%
\thanks{$^{2}$\,Sören Hohmann is with the Institute of Control Systems of the Karlsruhe Institute of Technology,
        76131 Karlsruhe, Germany
        \texttt{soeren.hohmann@kit.edu}}%
}

\maketitle
\thispagestyle{empty}
\pagestyle{empty}

\begin{abstract}
Robust and performant controllers are essential for industrial applications.
However, deriving controller parameters for complex and nonlinear systems is challenging and time-consuming.
To facilitate automatic controller parametrization, this work presents a novel approach using deep reinforcement learning (DRL) with N-dimensional B-spline geometries (BSGs).
We focus on the control of parameter-variant systems, a class of systems with complex behavior which depends on the operating conditions.
For this system class, gain-scheduling control structures are widely used in applications across industries due to well-known design principles.
Facilitating the expensive controller parametrization task regarding these control structures, we deploy an DRL agent.
Based on control system observations, the agent autonomously decides how to adapt the controller parameters.
We make the adaptation process more efficient by introducing BSGs to map the controller parameters which may depend on numerous operating conditions.
To preprocess time-series data and extract a fixed-length feature vector, we use a long short-term memory (LSTM) neural networks.
Furthermore, this work contributes actor regularizations that are relevant to real-world environments which differ from training.
Accordingly, we apply dropout layer normalization to the actor and critic networks of the truncated quantile critic (TQC) algorithm.
To show our approach's working principle and effectiveness, we train and evaluate the DRL agent on the parametrization task of an industrial control structure with parameter lookup tables.
\end{abstract}

\section{Introduction}
\label{sec:intro}
In industrial domains and applications, modern control systems safely operate complex systems with high tracking performance.
This is achieved by carefully applying approaches from linear and nonlinear control theory, sophisticated control system design, and the interplay of system identification, model calibration, and controller parametrization \cite{Rugh.2000, Huang.2022}.
For system dynamics that depends on the operating conditions, gain-scheduling-like control laws are a widespread choice~\cite{Rugh.2000}, yielding stable and robust performance over an operating range, e.g., parametrized with \ac{LMI} methods \cite{Huang.2022}.
However, with increasing control system complexity, the controller parametrization for these control structures becomes an expensive engineering task \cite{Dogru.2022, Kopf.2020}.

\begin{figure}[!t]
    \vspace{0.5em}
    \centering
    \includegraphics[width=0.99\linewidth]{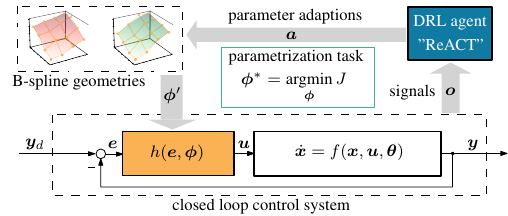}
    \vspace{-1.5em}
    \caption{We propose ReACT to effectively learn parametrization strategies for fixed control structures that depend on operating conditions. Our DRL agent observes closed-loop signals $\bm{o}$ and acts through parameter adaptations $\bm{a}$ using B-spline geometries (BSGs). ReACT improves the controller parameters $\bm{\phi}$  toward the control performance objective $J$.}
    \label{fig:paper_summary}
    \vspace{-1.5em}
\end{figure}

In recent years, autonomous approaches with \ac{RL} agents have been researched to compete with increasing complications by learning a parametrization processes for industrial control.
In \cite{Shipman.2019}, an actor-critic \ac{DRL} algorithm is applied on time-series observations as inputs to a feed-forward neural network.
The authors compound the reward function during training and concern randomized plant parameters of fixed system dynamics and noisy measurements.
Whereas \cite{Dogru.2022} focuses on process control with varying parameters under safe exploration of the parameter space.
However, reference tracking is not explicitly regarded and pre-tuning by an industrial controller is required.
In \cite{McClement.2022a}, the authors address varying parameters by implicitly learning meta-dynamics of individual systems with similar dynamics.
Though, initial knowledge of the dominant system parameters is required and the limitations regarding the sample efficiency under a wide distribution of system dynamics are discussed.
Recently, an agent with recurrent \ac{LSTM} neural networks for optimizing entire parameter lookup tables on closed-loop test-cases for the thermal management control system of a real car has been proposed \cite{Muhl.2022}.

In summary, the above works provide suitable methods but leave a gap toward autonomous \ac{DRL} agents for fixed-structure controllers with a priori unknown system dynamics and spatial parameter dependencies.
Additionally, there is a need for a capable action-space and agent design supporting sample efficiency, generalization, and stable training.

To close this gap, we propose a learning controller parametrization approach based on \acp{BSG}, as illustrated in Fig.~\ref{fig:paper_summary}.
We strive toward robust and efficient \ac{RL} policies to solve the expensive parametrization task with reference tracking by gain-scheduling controllers.
Our \ac{DRL} approach optimizes the closed-loop control performance $\bm{J}$ across operating ranges by adapting the \ac{BSG}-based controller parameters $\bm{\phi}$ from which parameter lookup tables for industrial implementations can be derived.
The proposed \ac{ReACT} framework autonomously operates on the closed-loop system without knowledge about underlying varying system parameters $\bm{\theta}$.

The main contributions of this work are: 1) The efficient parametrization of high-dimensional parameter spaces with \acp{BSG} as the interface to the \ac{RL} action space, 2) the \ac{RL} problem design that incorporates a self-competition reward, and 3) our regularized \ac{ReACT} agent with ablations on state-of-the-art off-policy algorithms, also concerning the actor neural network through dropouts and layer normalization for improved training under high variance.

In the following Section \ref{sec:background}, we formulate the parametrization problem and state preliminaries about \ac{RL} and \acp{BSG}.
In Section \ref{sec:approach}, we propose the novel \ac{DRL} agent design \ac{ReACT} and describe the training procedure.
We demonstrate the approach's effectiveness on exemplary parameter-varying system dynamics under varying dead time in Section \ref{sec:experiment}.

\section{Background}
\label{sec:background}
In the following, we define the controller parametrization task as the underlying problem we aim to solve.
We briefly introduce \ac{RL} as a suitable method for our extended B-splines solution approach.
Accordingly, the properties of B-splines and their flexibility in a parameter spaces, which are spanned across their dependencies, are pointed out.

\subsection{Problem Formulation}
We regard the family of nonlinear parameter-varying dynamic systems in the form
\begin{equation}\label{eq:lpv}
	\dot{\bm{x}}(t) =
	\bm{f} \left(
	\bm{x}(t), \bm{u}(t), \bm{\theta}(\bm{x}, \bm{u}, \bm{w})
	\bm \right),
\end{equation}
with system states $\bm{x}$, inputs $\bm{u}$, operating point dependent parameters $\bm{\theta}$ as well as exogenous inputs $\bm{w}$, see \cite{Rugh.2000}.

Widely established controller structures for this system class compensate for the influence of changing parameters on the system dynamics through adaptive controller gains.
Therefore, we assume the gain-scheduling control law
\begin{equation}\label{eq:gain_scheduling}
	\bm{u}(t) = \bm{g}\left(\bm{x}(t), \bm{y}(t), \bm{w}(t), \bm{\phi}(\bm{x}, \bm{y}, \bm{w}) \right)
\end{equation}
as the predefined and fixed control design \cite{Rugh.2000}.
The function $\bm{g}$ comprises the system states $\bm{x}$, the system outputs $\bm{y}$, and the mapping functions $\bm{\phi}$.
In every time step, the parameterized and dependent mapping $\bm{\phi}$ evaluates the current operating conditions ($\bm{x}, \bm{y}, \bm{w}$) for gain values which thereby creates an adaptive control law.
In industry applications of fixed structure controllers, the parameters $\bm{\phi}$ are often implemented as multidimensional lookup tables \cite{Rugh.2000, Muhl.2022}.

The parametrization task is given as the optimization problem to find the optimal parameter mapping
\begin{equation}\label{eq:optimization_problem}
	\bm{\phi}^* = \underset{\bm{\phi}}{\mathrm{argmin}}\,J
\end{equation}
that minimizes a control objective function $J \in \mathbb{R}^+$.
Commonly, a quadratic cost function
\begin{equation}\label{eq:lqr}
    J = \,\sum_{t_i=0}^{T-1} \bm{x}_{t_i} ^\mathrm{T} \bm{Q} \bm{x}_{t_i} + \bm{u}_{t_i} ^\mathrm{T} \bm{R} \bm{u}_{t_i}
\end{equation}
is utilized along a discrete time window of length $T$ and with discrete time steps $t_i$.
The cost function $J$ regards the system states $\bm{x}$, the inputs $\bm{u}$, and the positive weighting matrices $\bm{Q} \geq \bm{0}$ and $\bm{R} \geq \bm{0}$.
This paper focuses on the unconstrained case, referring to existing extensions for constrains \cite{Dogru.2022}.

We assume the availability of sufficient information about the closed-loop tracking control and the system dynamics through measurements.
Further, we regard the nonlinear system dynamics $\bm{f}$ as unknown in detail, which is mostly the case across applications in industry \cite{Rugh.2000}.
In accordance with the practical design, we regard the controller parametrization task as a sequential procedure.
The parameter mappings $\bm{\phi}$ are iteratively adapted after the control performance is assessed in fixed interval.

\subsection{Reinforcement Learning}
With increased system dynamics complexity, or degrees of freedom in the controller, as well as the absence of an exact system model, the formulated problem in (\ref{eq:optimization_problem}) quickly becomes infeasible to solve with traditional methods, such as nonlinear or dynamic programming.

The described parameter adaptation process is regarded as a sequential decision-making problem that can approximately be modeled as an \ac{MDP} and addressed with \ac{RL} \cite{Sutton.2018}.
Accordingly, the measurable information about the system is assumed as sufficient to solve this \ac{MDP} with an \ac{RL} agent.
An \ac{RL} agent adapts the controller parameters $\bm{\phi}$ with its actions $\bm{a}$ which are based on the observations $\bm{o}$.
A rewarding feedback $r$ enables the agent to learn which adaptations are expected to result in an improved control performance with respect to the objective $J$.
We consider \ac{DRL} as a promising solution concept to learn a strategy efficiently solving the controller parametrization task in (\ref{eq:optimization_problem}) \cite{McClement.2022a, Muhl.2022}.
Therefore, we regard the closed-loop control system with its parameter tuning task using B-spline geometries in Fig.~\ref{fig:paper_summary} as the \ac{RL} environment.

A control parameter optimization for $J$ relaxes in an \ac{RL} formulation to the optimization of the expected sum of discounted future rewards \cite{Sutton.2018}:
\begin{equation}
    \underset{\bm{\pi}}{\text{argmax}}
    E \left\{
        \sum_{k} \gamma^k r_k
    \right\},
\end{equation}
with discounting factor $\gamma \in \left[ 0, 1 \right]$ and the \ac{RL} policy $\bm{\pi}$.
For our design, we deploy actor-critic algorithms that also learn to approximate a state-action $Q$-value \cite{Sutton.2018}.

Furthermore, off-policy model-free algorithms, i.e., \ac{SAC} \cite{Haarnoja.2018} and its derivatives DroQ \cite{Hiraoka.2022}, and \ac{TQC} \cite{ArseniiKuznetsov.2020}, are respected to improve the sample efficiency limitations (cf. on-policy \ac{RL} in \cite{McClement.2022a}).

\subsection{B-spline Geometries}

\begin{figure}[!t]
    \centering
    \includegraphics[width=0.99\linewidth]{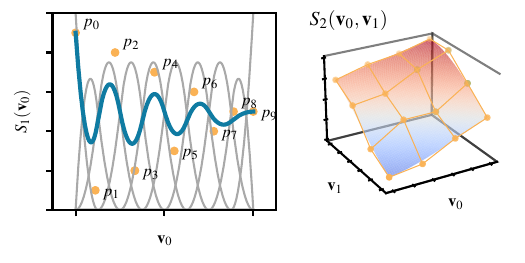}
    \vspace{-2em}
    \caption{An exemplary oscillating curve (left, blue)   approximated with a univariate \ac{BSG}-curve $S_1(\bm{\mathbf{v}}_0)$ through its basis functions (gray) and a BSG-surface $S_2(\bm{\mathrm{v}}_0, \bm{\mathrm{v}}_1)$ spanned with a bivariate \ac{BSG} (right). The CPs $\bm{p}_{i_n}$ (yellow dots) shape the geometries accordingly.}
    \label{fig:b_splines}
    \vspace{-1.5em}
\end{figure}

For the representation and adaptation of high-dimensional continuous parameter spaces, we regard \acp{BSG} in this work.
B-splines are suitable due to their continuous differentiability, partitioning into parameter sub-spaces, and computational efficiency.

The $N$-dimensional \ac{BSG} $S_N$ is defined as the weighted product of B-spline basis functions $b_{i_n}(\bm{\mathrm{v}}_n)$ \cite{GeraldFarin.2002}:
\begin{equation}
    S_N\left(\bm{\mathrm{v}}_0, \ldots, \bm{\mathrm{v}}_{N-1}\right)
    =
    \sum_{i_0=1}^{M_0} \dots \sum_{i_{N-1}=1}^{M_{N-1}}
    p_{i_0 \dots i_{N-1}}
    \prod_{n=0}^{N-1}
    b_{i_n}\left(\bm{\mathrm{v}}_n\right),
\end{equation}
where $p_{i_0 \dots i_{N-1}}$ are the \acp{CP} with a cardinality ${M_n = |\bm{p}_{i_n}|}$ in the $n$-th dimension of the \ac{BSG}.
Each dimension $n$ is described with $M_n$ B-spline basis functions $b_{i_n, d}\left(\mathrm{v}_n\right)$ of polynomial degree $d$.
They are piecewise defined and based on the knot vector $\bm{\mathrm{v}}_{n} = [\mathrm{v}_{0_n}, \mathrm{v}_{1_n}, ..., \mathrm{v}_{(M_n+d+1)_n}]$ with knots $\mathrm{v}_{i_n}$ \cite{LesPiegl.1996}.
For simplicity, we state $\mathrm{v}_{i_n} = \mathrm{v}_{i}$ for the definition of the B-spline basis functions along one axis:
\begin{align}
    b_{i, 0}(\mathrm{v})  =& \left\{
        \begin{array}{cl}
            1 & \text { if } \mathrm{v}_{i} \leq \mathrm{v}<\mathrm{v}_{i + 1} \\
            0 & \text { otherwise }
        \end{array}\right.
\end{align}
in the case of degree $d = 0$, and for $d > 0$ as
\begin{align}
    b_{i, d}(\mathrm{v}) = &
        \frac{\mathrm{v}-\mathrm{v}_{i}}{\mathrm{v}_{i+d} - \mathrm{v}_{i}} b_{i, d-1}(\mathrm{v})
        \\ & +
        \frac{\mathrm{v}_{i+d+1} - \mathrm{v}}{\mathrm{v}_{i + d + 1}-\mathrm{v}_{i+1}} b_{i+1, d-1}(\mathrm{v}).
    \nonumber
\end{align}

Univariate \ac{BSG}-curves ($N=1$) and bivariate ($N=2$) \ac{BSG}-surfaces are widely used and visualized in Fig.~\ref{fig:b_splines}, respectively.
The univariate case is depicted with the yellow \acp{CP} $p$, the grey B-spline basis functions, and the resulting curve in blue.
To account for higher dimensional dependencies, \acp{BSG} with ${N\ge3}$ can be used \cite{Tustison.2006}.
We apply \acp{BSG} as a suitable mathematical tool to represent high-dimensional controller parameters with parameter-variant system dependencies.
Moreover, the \acp{CP} $p$ support the effective adaptation of parameter spaces.
For further insights into the general theory of \ac{BSG}, we refer to \cite{GeraldFarin.2002} and \cite{LesPiegl.1996}.
Further kernels that match application-specific requirements can be used as well, e.g., the related \acp{NURBS} \cite{LesPiegl.1996} or radial basis functions \cite{Cao.2020}.

\section{Approach}\label{sec:approach}

In this section, we present our novel \ac{DRL} approach for efficient and robust \ac{BSG}-based controller parametrization.
We define the \ac{RL} environment $\mathcal{E}$ and show the use of the B-spline \acp{CP} to reduce the agent's action space.
Subsequently, we describe the general training algorithm of the \ac{DRL} agent to solve the controller parametrization task.
Finally, we discuss suitable regularization techniques and present our \ac{DRL} agent structure (\ac{ReACT}), which improves the agent's learning and inference behavior for a successful transition toward real-world applications.

\begin{figure*}[!t]
    \centering
    \vspace{0.5em}
    \includegraphics[width=0.99\linewidth]{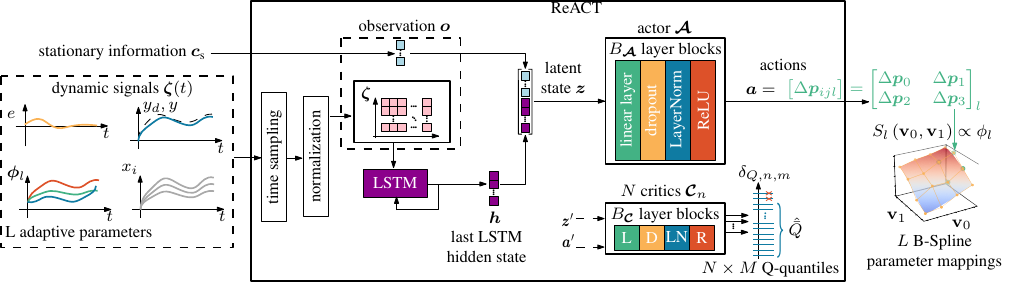}
    \caption{Structure of the proposed ReACT agent which observes closed-loop time series data $\bm{\zeta}(t)$ concatenated with stationary information $\bm{c}_\text{s}$.
    The extracted feature vector $\bm{z}$ represents the latent environment state.
    The agent's actions $\bm{a}$ incrementally adjust the B-spline CPs of controller parameters $\phi_l$.}
    \label{fig:agent_architecture}
    \vspace{-1em}
\end{figure*}

The \ac{RL} environment $\mathcal{E}$ comprises the closed-loop control system and the parameter adaptations as shown in Fig.~\ref{fig:paper_summary}.
The interaction of the \ac{ReACT} agent with the environment $\mathcal{E}$ and the agent's detailed structure is illustrated in Fig.~\ref{fig:agent_architecture}. By interacting with $\mathcal{E}$, the agent receives observations $\bm{o}$ that provide sufficient information about the control system behavior.
We regard stationary information $\bm{c}_s$ and transient signals $\bm{\zeta}(t)$.
The stationary information $\bm{c}_s$ remains constant for an observed time window, e.g., like slowly changing states of the system.
In contrast, the time-varying signals $\bm{\zeta}(t)$ dynamically change with respect to the system dynamics in (\ref{eq:lpv}) and (\ref{eq:gain_scheduling}).
Accordingly, we consider the observation
\begin{align}\label{eq:observation}
    \bm{o} &= \left[ \bm{c_s}, \bm{\zeta}(t) \right] \\
        &= \left[ \bm{c_s}, \bm{y}_\mathrm{d}(t), \bm{y}(t), \bm{e}(t), \bm{\hat{x}}(t), \bm{u}(t), \bm{w}(t) \right],
\end{align}
with system outputs $\bm{y}$, desired references $\bm{y}_\text{d}(t)$, tracking control error $\bm{e}(t) = \bm{y}_\text{d}(t) - \bm{y}(t)$, and estimated system states $\hat{\bm{x}}(t)$ if regarded as relevant information for the individual \ac{BSG}-based controller parametrization task.

Based on a given observation $\bm{o}$, the agent derives the next action $\bm{a}$, shown on the right side of Fig.~\ref{fig:agent_architecture}.
To reduce the parameter search-space dimension, we use \ac{BSG}-kernels $S_l$ which efficiently approximate the controller parameter mappings $\phi_l$.
We regard the changes to the \acp{CP} $\bm{p}_{i_n}$ as actions $\bm{a}$ of the proposed \ac{ReACT} agent:
\begin{align}\label{eq:action}
    \bm{a} &= \Delta \bm{p}_{i_n}.
\end{align}
Accordingly, we adjust the \acp{CP} values at step $k$ by
\begin{equation}
    \bm{p}_{i_n; k} = \bm{p}_{i_n; k-1} + \Delta \bm{p}_{i_n; k}.
\end{equation}
Due to the beneficial property of B-splines, \acp{BSG} are only modified in a local area when changing a limited subset of the \acp{CP}.
For embedded applications, controller parameter lookup tables can be retrieved in a favorable discretization through pointwise evaluation of the \acp{BSG}.
Our approach supports $L$-multiple \acp{BSG} $S_l$ to represent the n-dimensional controller parameters $\phi_l$, as visualized in Fig.~\ref{fig:agent_architecture}.

On the update of the controller parameters $\bm{\phi}$, we reward the agent based on the control objective with $r_J=-J$, see~\eqref{eq:lqr}.
However, the design of a suitable reward function $r$ for the given parametrization task is challenging due to the variance in the system dynamics and the complex controller parameter space.
In consequence, we propose to regularize the training process with additional reward terms to facilitate convergence.
We discourage parameter oscillations by penalizing unnecessary actions through $r_a = - \nicefrac{1}{N} \sum_n | a_n |$ \cite{McClement.2022a} and apply a self-competition reward \cite{Mandhane.2022}:
\begin{equation}\label{eq:self_competition_reward}
    r_\text{sc} =
    \begin{cases}
        \text{1 if } r_J > \text{EMA}(\bar{r}_J) \\
        0 \text{ else}
    \end{cases}
    ,
\end{equation}
with the \ac{EMA} of prior episodes' mean control performances $\bar{r}_J$.
This encourages the \ac{ReACT} agent to improve continuously.
In summary, the reward function is defined as
\begin{equation}\label{eq:reward_function}
r(\bm{o}, \bm{a}) = b_1 \cdot r_J + b_2 \cdot r_{\bm{a}} + b_3 \cdot r_\text{sc},
\end{equation}
with weightings $\left\{ b_1, b_2, b_3 \right\} \in \mathbb{R}^+$ as design choices.

Since our proposed \ac{ReACT} approach operates on discrete-step decisions and leverages neural networks for function approximations, time series of the dynamic signals $\bm{\zeta}(t)$ are first time sampled with $\Delta t>0$ and then normalized to suitable signal value ranges.
Subsequently, an \ac{LSTM} neural network processes the sequence and feeds back its current outputs $\bm{h}$ for each signal time sample $t_i$, see \ref{eq:lqr}.
The last hidden state $\bm{h}$ of the \ac{LSTM} contains learned features about the input signals which are then concatenated with the stationary information $\bm{c}_s$ to form a latent state $\bm{z}$.

\begin{algorithm}[!b]
    \caption{Training routine for the \ac{BSG}-based controller parametrization}\label{alg:rl_training}
    \begin{algorithmic}[1]
        \STATE \textbf{input} training trajectory set $\mathcal{Y}_d$, system parameter dataset $\mathcal{D}_{\bm{\theta}}$, control parameter \acsp{BSG} $S_l$
        \STATE \textbf{initialize} \ac{DRL} agent policy $\bm{\pi}_{\bm{\psi}}$, Q-networks $\bm{Q}_{\bm{\omega}}$, experience replay buffer $\mathcal{R}$, max. episode steps $K$
        \FOR{each episode}
            \STATE Sample $\bm{\theta}$ from $\mathcal{D}_{\bm{\theta}}$ %
            \STATE $\bm{p}_{i_n l} \gets \bm{p}_{0; i_n l}$ %
            \STATE $S_l\left(\mathrm{v}_0, \ldots, \mathrm{v}_N\right) \gets $ $ \bm{p}_{i_n l}$ %
            \STATE $\phi_{0;l} \gets $ Evaluate($ S_l\left(\mathrm{v}_0, \ldots, \mathrm{v}_N\right)$) %
            \STATE Sample $\bm{y}_d$ from $\mathcal{Y}_d$ %
            \STATE $\bm{o_0} \gets \left[ \bm{c_s}, \bm{\zeta}(t) \right] $ %
            \STATE $k \gets 1$
            \WHILE{$k \leq K$}
                \STATE $\bm{a} \gets \bm{\pi}_{\bm{\psi}}(\bm{o})$ %
                \STATE $\bm{p}_{i_n l} \gets $ Apply($  \bm{a}_k $)  %
                \STATE $S_l\left(\mathrm{v}_0, \ldots, \mathrm{v}_N\right) \gets \bm{p}_{i_n l} $
                \STATE $\phi_l \gets $ Evaluate($ S_l\left(\mathrm{v}_0, \ldots, \mathrm{v}_N\right)$)
                \STATE Sample $\bm{y}_d$ from $\mathcal{Y}_d$ %
                \STATE $\bm{o}_{k+1} \gets \left[ \bm{c_s}, \bm{\zeta}(t) \right] $
                \STATE $r_{k} \gets $ CalculateRewards($\bm{o}_k, \bm{a}_{k}, \bm{o}_{k+1}) $
                \STATE Store $(\bm{o}_{k}, \bm{a}_{k}, r_k, \bm{o}_{k+1})$ in $\mathcal{R}$
                \STATE Update Q-networks $\bm{Q}_{\bm{\omega}}$
                \STATE Update policy network $\bm{\pi}_{\bm{\psi}}$
                    \IF{$k \geq$ $K$}
                \STATE End episode
                \ENDIF
                \STATE $k \gets k + 1$
            \ENDWHILE
        \ENDFOR
        \STATE \textbf{output} agent policy $\bm{\pi}_{\bm{\psi}}$
    \end{algorithmic}
\end{algorithm}

The acting policy $\bm{\pi}$ (actor $\bm{\mathcal{A}}$) as well as the critics $\bm{\mathcal{C}}_n$, which estimate the state-action values $Q$, are based on the latent vector $\bm{z}$ input.
To reduce the overestimation of $Q$-values and stabilize learning, we apply the \ac{TQC} algorithm for our \ac{BSG}-based controller parametrization approach.
The \ac{TQC} critics comprise an ensemble of $Q$-estimators to create a mixture of experts through quantile regression~\cite{ArseniiKuznetsov.2020}.
The networks $\bm{\pi}_{\bm{\psi}}$ (actor $\mathcal{A}$) and $\bm{Q}_{\bm{\omega}}$ (critics $\mathcal{C}_n$) are parametrized by $\bm{\psi}$ and $\bm{\omega}$, and trained with backpropagation of the \ac{TQC} policy and Q objective gradients~\cite{ArseniiKuznetsov.2020}.

In \cite{Hiraoka.2022}, dropout and layer-normalization as regularizations enable multiple $Q$-network updates per environment step to further increase the sample efficiency, meaning the number of costly environment steps until learning converges.
We regard the regularization techniques proposed in \cite{Hiraoka.2022} and \cite{ArseniiKuznetsov.2020} as orthogonal concepts which we extend to uncertainty-affected controller parametrization tasks for our \ac{ReACT} agent.
We propose to apply the \ac{DroQ} network regularization from \cite{Hiraoka.2022} by dropout and layer normalization (yellow and blue layers) to the actor $\mathcal{A}$, as depicted in Fig.~\ref{fig:agent_architecture}.

In summary, the above leads to the novel agent structure \acf{ReACT} and offers the advantage of more robust policies for closed-loop system disturbances.

The adapted training routine for the \ac{BSG}-based controller parametrization task is described in Algorithm~\ref{alg:rl_training}.
It requires a given training reference trajectory set $\mathcal{Y}_d$ and a system definition dataset $\mathcal{D}_{\bm{\theta}}$ with the system parameters $\bm{\theta}$.
First, the experience replay buffer $\mathcal{R}$ storing the \ac{RL} experience tuples for a batch-wise update of the agent's networks, the networks ($\bm{\pi}_{\bm{\psi}}$ and $\bm{Q}_{\bm{\omega}}$), and the maximum episode steps $K$ are initialized.
For each episode, the system dynamics with mapping $\bm{\theta}(\bm{x},\bm{y},\bm{w})$ is kept unchanged, shown in line 4, and the new control system is observed.
The training procedure, then considers a fixed number of steps $K$ per episode.
For every step $k$, one training trajectory tuple $\bm{y}_d$ from $\mathcal{Y}_d$ is chosen as described in lines 8 and 16.

To ensure an efficient training procedure, the training trajectories $\bm{y}_d \subseteq \mathcal{Y}_d$ should sufficiently cover a possible operating range and the underlying system parameter dependencies.
After the successful episodic training from updates of the \ac{ReACT} agent in lines 17-21, its policy network $\bm{\pi}(\bm{o})$ can be inferred with closed-loop control system observations $\bm{o}$ to adjust the controller parameters $\bm{\phi}$ automatically.

\section{Experiment Results}\label{sec:experiment}
\label{sec:results}
To demonstrate the effectiveness of the proposed \ac{ReACT} agent, we design an experimental setup with a parameter-variant system inspired by industrial use cases that demand significant control system calibration and extensive parametrization effort.
We describe the training procedure and conduct an ablation study concerning competing \ac{DRL} algorithms and the additional actor regularization.
Furthermore, we compare the \ac{ReACT} based parametrization to a base controller parametrization and discuss the control performance and robustness of the closed-loop control system.

\begin{figure}[!b]
    \vspace{-1em}
    \centering
    \includegraphics[width=0.99\linewidth]{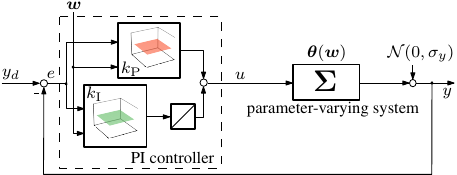}
    \caption{Exemplary closed-loop PI control structure with the parameter-varying system $\bm{\Sigma}$ and gain-scheduling lookup tables $\bm{\phi} = \left\{ k_P, k_I \right\}$.}
    \label{fig:control_system}
\end{figure}

\subsection{Experiment Setup}\label{subsec:experiment_setup}

The experimental setup, as described in Fig.~\ref{fig:control_system}, focuses on a gain-scheduling controlled parameter-variant system $\bm{\Sigma}$, with unknown dynamic parameters $\bm{\theta}(\bm{w})=[\theta_1(\bm{w}), \theta_2(\bm{w})]^\text{T}$ that depend on exogenous inputs $\bm{w}$.
The exogenous inputs $\bm{w}=[w_1, w_2]^\text{T}$ refer to operating points during dynamic operation.
The controlled plant is modeled as parameter-varying \ac{FOPDT} dynamics
\begin{equation}
    \bm{\Sigma}:\quad
    \dot{x}(t) =
    - \frac{1}{\theta_1(\bm{w})} \cdot \left(
        x(t)
        + u(t - \theta_2(\bm{w}))
    \right).
\label{eq:fopdt}
\end{equation}
The system dynamics represent an extension of the test system in \cite{McClement.2022a}, where fixed parameters in the \ac{FOPDT} dynamics are applied.
Here, $\theta_1$ represents the transient-time parameter, and $\theta_2$ denotes the dead-time parameter.
The system output $y$ and operating points $\bm{w}$ are disturbed with additive zero-mean Gaussian noise to account for real-world applications.

We apply a \ac{PI} control law for tracking a desired output $y_d$ as
\begin{equation}
    u(t) = k_\text{P}(\bm{w}) \cdot e(t) + k_\text{I}(\bm{w}) \cdot \int e(\tau) d\tau,
\end{equation}
with the proportional and integral gains $k_\text{P}$ and $k_\text{I}$ with two-dimensional dependencies on the exogenous inputs $\bm{w}$. They act as scheduling variables to address the external influences on the system dynamics with the controller.

\begin{figure}[!b]
    \centering
    \resizebox{1.02\linewidth}{!}{
        \input{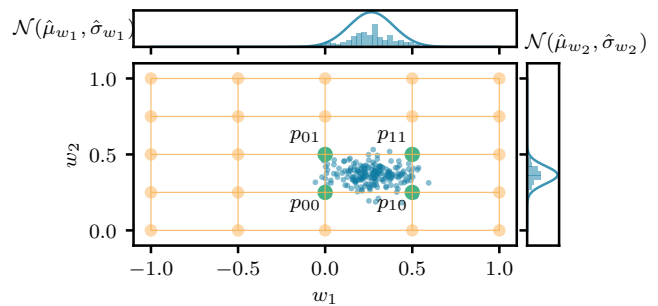}
    }
    \vspace{-2em}
    \caption{BSG-surface CPs in yellow with selected CPs in green which are adjusted by our ReACT agent. The selection is defined around the mean and the standard deviation along each of the dependencies $\bm{w}$.}
    \label{fig:actions}
\end{figure}

The training is conducted on a closed-loop control system simulation, implemented as a C++ code generated \matlab \simulink model.
It is interfaced with C++/Python bindings in an \ac{RL} \textit{Gym} environment. %
We incorporate the proposed \ac{ReACT} agent and its architecture, as shown in Fig.~\ref{fig:agent_architecture}, into the \ac{DRL} framework \textit{Stable Baselines3} \cite{Raffin.2021}.

\begin{figure*}[!ht]
    \centering
    \vspace{0.25em}
    \resizebox{1.00\textwidth}{!}{
        \input{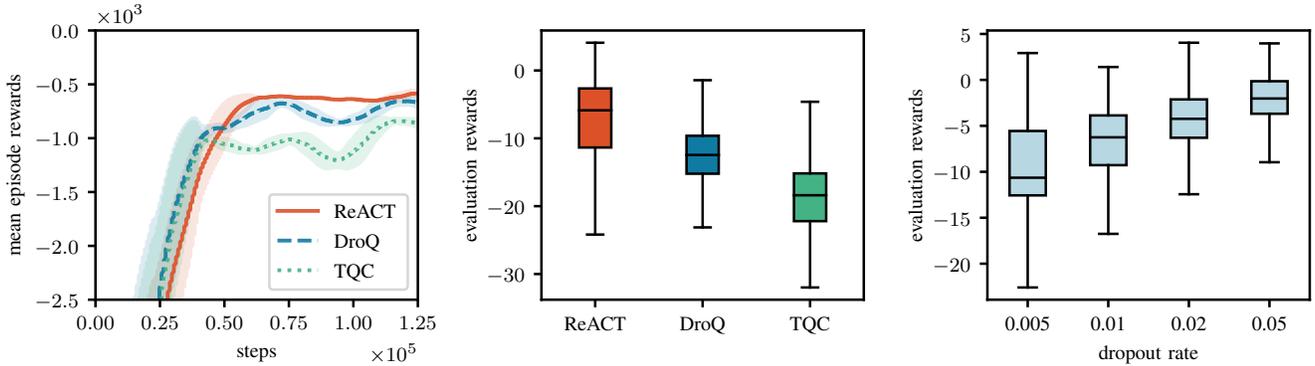}
    }
    \caption{Comparison of our ReACT agent with RL baseline algorithms: training convergence (left) and task performance of the learned policies (center). An ablation study on regularizing dropouts (right) shows the positive influence of actor dropout rates for the controller parametrization task.}
    \label{fig:reward_curves}
    \vspace{-1em}
\end{figure*}

We approximate the controller parameter spaces by $\bm{\phi} = \left\{ \phi_\text{P},\phi_\text{I} \right\}$ using \ac{BSG}-surfaces $S_{k_\text{P}}$ and $S_{k_\text{I}}$ with the $2$-dimensional \ac{CP} grids $\bm{p}_{k_\mathrm{P}}\left( \bm{\mathrm{v}}_0, \bm{\mathrm{v}}_1 \right) \text{ and } \bm{p}_{k_\mathrm{I}}\left( \bm{\mathrm{v}}_0,\bm{\mathrm{v}}_1 \right)$.
Our \ac{ReACT} agent adapts through its actions the proportional and integral controller gains $\left\{ k_\text{P}, k_\text{I} \right\}$ via these \ac{CP} grids of the \ac{BSG}-surfaces.
In order to simulate and embedded application, the controller parameters are implemented as lookup tables which are sampled from evaluating the \ac{BSG}-surface.
A controller grid's size leads to a trade-off between a well-approximated parameter space, parametrization effort, and control performance.
Specifically, a high-resolution controller grid is required for good control behavior if sensitive dependencies on the operating points exist.
The degree $d$ of the basis functions is a design choice to regulate the influence of the control points on the surface.

The task observation space comprises eight time series
\begin{equation}
    \bm{o} =
        \begin{bmatrix}
        y_d(t) &
        y(t) &
        e(t) &
        k_\text{P}(t) &
        k_\text{I}(t) &
        u(t) &
        \bm{w}(t)
        \end{bmatrix}.
\end{equation}
The agent's action space is a fixed $8$-dimensional vector
\begin{align}
    \bm{a} &= \left[ \bm{a}_{k_\mathrm{P}}, \bm{a}_{k_\mathrm{I}} \right] \\
    &=
    \begin{bmatrix}
        \Delta p_{k_\mathrm{P},\,0} & \Delta p_{k_\mathrm{P},\,1} & \Delta p_{k_\mathrm{P},\,2} & \Delta p_{k_\mathrm{P},\,3} \\
        \Delta p_{k_\mathrm{I},\,0} & \Delta p_{k_\mathrm{I},\,1} & \Delta p_{k_\mathrm{I},\,2} & \Delta p_{k_\mathrm{I},\,3}
    \end{bmatrix}^T,
\end{align}
where $\bm{a}_{k_\mathrm{P}}$ and $\bm{a}_{k_\mathrm{I}}$ are applied to a reduced set of the available \acp{CP} of
$\bm{p}_{k_\mathrm{P}}\left( \bm{\mathrm{v}}_0, \bm{\mathrm{v}}_1 \right)$ and $\bm{p}_{k_\mathrm{I}}\left( \bm{\mathrm{v}}_0, \bm{\mathrm{v}}_1 \right)$, respectively.
We determine a region of interest based on the mean and the standard deviation of the operating points trajectories $w_1(t)$ and $w_2(t)$ along the time window.
Accordingly, we select the next outer \acp{CP} along each operating point dependency axis as illustrated in Fig.~\ref{fig:actions}.
The agent's actions $\bm{a}$ are applied on these \acp{CP} marked in green.

We reward the task performance via the quadratic control objective~\eqref{eq:lqr} with the control deviation $e$ and control $u$:
\begin{equation}
    r_J =
    - \frac{1}{T} \sum_{t_k=0}^{T-1} \left( Q \cdot e^2(t_k) + R \cdot u^2(t_k) \right) \cdot \Delta t,
\end{equation}
with the simulation trajectory length $T$, the discrete sample time ${\Delta t=\SI{50}{\milli\second}}$, and the design choice $Q=0.99$ and $R=0.01$.
We set the weights of the shaped reward function~\eqref{eq:reward_function} to $b_1=100$ and $b_2=b_3=1$ which prioritizes the main task of parametrization in regard to the control objective.
The regularization term weights $b_2$ and $b_3$ can be optimized for training convergence speed.

For the training routine, the input training trajectory set $\mathcal{Y}_d$ consists of step functions with random-uniform sampled steps, operating points and initial condition.
The system parameter set $\mathcal{D}_{\bm{\theta}}$ comprises functions of $\bm{\theta}$, depending on $\bm{w}$, see \eqref{eq:fopdt}.
During training, we sample new test trajectories $y_d$ at every episode step $k$, as given in lines 8 and 16 in Algorithm \ref{alg:rl_training}.
At the beginning of each new episode, we reset the \ac{BSG}-based controller parametrization by setting the default values for the \ac{CP} grids $\bm{p}_{k_\mathrm{P}}\left( \bm{\mathrm{v}}_0, \bm{\mathrm{v}}_1 \right)$ and $\bm{p}_{k_\mathrm{I}}\left( \bm{\mathrm{v}}_0, \bm{\mathrm{v}}_1 \right)$, representing a slow but stable parametrization baseline.

\subsection{Results}\label{subsec:results}

In the simulation study, we evaluate our \ac{BSG}-based controller parametrization approach using the given experimental setup.
The agent's hyperparameters are provided in Table~\ref{tab:experiment_parameters}.
To derive conclusive results concerning the designed control parametrization environment and the autonomous \ac{DRL} parametrization agents, we train and evaluate the model for $K=250$ steps per episode and average the results over 5 different random seeds.
Additionally, we compare our proposed \ac{ReACT} agent with the \ac{TQC} algorithm as its basis without regularizations, and with \ac{DroQ}, a critic-only regularized design.
The results demonstrate in Fig.~\ref{fig:reward_curves} that the proposed \ac{ReACT} outperforms the other agents in terms of rolling means of the episode rewards during training (depicted on the left) and during evaluation (depicted in the center of Fig.~\ref{fig:reward_curves}).
In particular, \ac{ReACT} exhibits less variance once it converges, indicating actor regularization with slightly slower but more consistent training.

Furthermore, we conduct a hyperparameter study for different dropout rates, illustrated in the right of Fig.~\ref{fig:reward_curves}, exposing that higher dropout rates lead to better evaluation rewards and less variance.
This indicates that the agent is more robust due to the included regularization.
However, with further increase of the dropout rates, we observe a performance saturation that is linked to the task maximum performance and capacities of the used neural network \cite{Hiraoka.2022}.

\begin{table}[!t]
    \centering
    \caption{DRL agent hyperparameters}
    \vspace{-0.5em}
    \label{tab:experiment_parameters}
    \begin{tabular}{l c}
        hyperparameter & value \\
        \hline
        optimizer & \texttt{Adam} \\
        number of samples per batch & \num{128} \\
        learning rate & \num{3e-4} \\
        gradient steps & \num{4} \\
        replay buffer size & \num{1e6} \\
        discount factor $\gamma$ & \num{0.90} \\
        entropy target & $-\dim(\bm{a})$\\
        entropy temperature factor & \num{1} \\
        target network smoothing coefficient ($\rho$) & \num{5e-3} \\
        number of LSTM layers & \num{2} \\
        number of hidden units per LSTM layer & \num{256} \\
        Q-networks & \num{2} \\
        number of blocks ($B_\mathcal{A}$, $B_\mathcal{C}$) & \num{2} \\
        number of hidden units per block & \num{256} \\
        activation functions & \texttt{ReLU} \\
        quantiles per Q-network & \num{25} \\
        top quantiles to drop per net & \num{2} \\
        dropout rates & \num{0.03} \\
        \hline
    \end{tabular}
    \vspace{-2em}
\end{table}

\begin{figure}[!t]
    \centering
    \resizebox{1.00\linewidth}{!}{
        \input{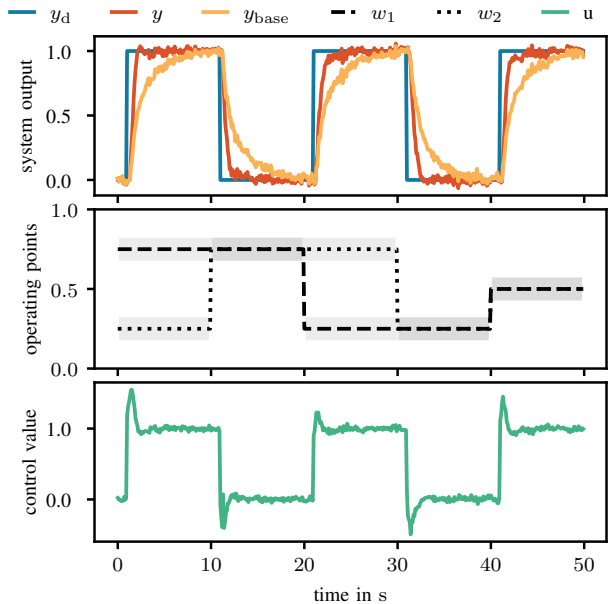}
    }
    \vspace{-2em}
    \caption{Performance of the parametrization baseline $y_\textrm{base}$ and the autonomous ReACT parametrization result $y$. The varying operating points $\bm{w}$ with standard deviations and the control $u$ are visualized.}
    \label{fig:simulation_results}
    \vspace{-1em}
\end{figure}

To evaluate the trained agent's parametrization capability, we deploy the trained policy $\bm{\pi}$ (actor $\bm{\mathcal{A}}$) to an exemplary system $\bm{\Sigma}$ from Fig.~\ref{fig:control_system} to adapt the controller parameters $k_P$ and $k_I$.
During policy rollout, we assess the performance based on operating-point-dependent step responses.
The results show that the policy efficiently adjusts the \ac{BSG} within 150 steps for a well-approximated controller parameter space.
Fig.~\ref{fig:simulation_results} presents a comparison of the control behavior between a stable base parametrization $y_{\mathrm{base}}$ and the adapted parametrization by the \ac{ReACT} agent.
The operating points $w_1$ and $w_2$ vary throughout the comparison to cover a broad range of the controller parameter space.
Additionally, the operating points are disturbed with zero mean Gaussian noise, depicted in Fig.~\ref{fig:simulation_results} with the standard deviation.
The \ac{ReACT} approach autonomously parametrizes the controller robustly with respect to the dead time and exhibits a faster control performance $y$ compared to the baseline $y_\text{base}$.

\subsection{Discussion}

Our \ac{ReACT} agent provides promising results that encourage ongoing studies on generalized controller parameter optimization for a range of system classes and toward real-world deployment.
One aspect that requires detailed validation is if the proposed regularized actor network design can be transferred into a robust application.

Another aspect is the stability of the closed-loop controlled system with the adapted controller parameters.
In \cite{Rugh.2000}, appropriate and well-known approaches regarding stability of gain-scheduling approaches are discussed.
Accordingly, traditional stability analysis can be applied on plant approximations together with the known fixed-structure control law.
This can be implemented as a verification step before setting the parameters.
However, the control system stability or robustness may already be considered during the training process.
This potentially results in a robust control behavior by design, e.g., via Lyapunov-based approaches, see \cite{Rudolf.2022}.

\section{Conclusion}

For complex system dynamics that depend on operating conditions, the controller parametrization task still demands significant engineering effort and time.
To facilitate an automatic and more efficient controller parametrization for n-dimensional lookup tables, we present a novel \ac{DRL} approach that includes B-spline geometries to approximate the controller parameter space.
The proposed \ac{ReACT} agent includes layer normalization and dropouts as regularizations for the actor and critic.
A benefit is shown in the training and agent inference under noise.
The learned strategy is utilized to incrementally adapt the controller parameters, resulting in optimized control performance.
We demonstrate the effectiveness of our approach by considering the controller parametrization of a parameter-varying \ac{FOPDT} system.

In this work, we limited the actions to pre-selected control points rather than learning them or using the partitioning properties of B-splines.
The latter can enable varying sensitivity of the parametrization over the operating range.
Our future work will leverage this flexibility and investigates toward further generalizing methods based on task context and meta-information about the parametrization task.

\addtolength{\textheight}{-20cm}   %

\vspace{1.1em}
\bibliographystyle{IEEEtran} %
\bibliography{indices/SMC_2023.zotero.bibtex}

\end{document}